\documentclass[conference]{IEEEtran}

\usepackage{microtype} 
\usepackage{booktabs} 
\usepackage{algpseudocode}
\usepackage{algorithm}
\usepackage{listings}
\usepackage{mathtools}
\usepackage{amsmath}
\usepackage{amssymb}
\usepackage{bm}
\usepackage{enumerate}
\usepackage[framemethod=TikZ]{mdframed}
\usepackage{xspace}
\usepackage[caption=false]{subfig}
\usepackage{cleveref}
\usepackage{relsize}
\usepackage[T1]{fontenc} 
\usepackage{color} 
\usepackage[hyphens]{url}
\usepackage{cite}
\usepackage{algorithm}
\usepackage{algpseudocode}
\usepackage{booktabs}
\usepackage{wrapfig}
\usepackage{balance}
\usepackage{multirow}
\usepackage[most]{tcolorbox}
\usepackage{xcolor}
\usepackage{array}
\usepackage{etoolbox}
\usepackage[normalem]{ulem}
\usepackage{soul}
\usepackage[english]{babel}

\definecolor{shadecolor}{RGB}{150,150,150}

\def\HiLi{\leavevmode\rlap{\hbox to \hsize{\color{black!20}\leaders\hrule height .8\baselineskip depth .5ex\hfill}}}

\graphicspath{{./figures/}}

\makeatletter
\newcommand{\algcolor}[2]{%
  \hskip-\ALG@thistlm\colorbox{#1}{\parbox{\dimexpr\linewidth-2\fboxsep}{\hskip\ALG@thistlm\relax #2}}%
}

\makeatother

\newcommand{\sfsmaller}{\smaller}
\newcommand{\bench}[1]{\textsf{\small #1}}
\newcommand{\heuristic}[1]{\textsf{\small #1}}
\newcommand{\kfusion}{KinectFusion\xspace}
\newcommand{\Kfusion}{KinectFusion\xspace}
\newcommand{\kinectfusion}{KinectFusion\xspace}
\newcommand{\KFusion}{KinectFusion\xspace}
\newcommand{\knob}[1]{\textsf{#1}}
\newcommand{\code}[1]{\texttt{#1}}

\newcommand{\eg}{e.g.\xspace}
\newcommand{\ie}{i.e.\xspace}

\makeatletter
\NewDocumentCommand{\LeftComment}{s m}{%
  \Statex \IfBooleanF{#1}{\hspace*{\ALG@thistlm}}\(\triangleright\) #2}
\makeatother

\tcbset{
    frame code={}
    center title,
    left=0pt,
    right=0pt,
    top=0pt,
    bottom=0pt,
    colback=gray!20,
    colframe=white,
    width=\dimexpr\textwidth\relax,
    enlarge left by=0mm,
    boxsep=5pt,
    arc=0pt,outer arc=0pt,
    }

\newcommand{\name}{SLAMBooster\xspace}

\newcolumntype{H}{>{\setbox0=\hbox\bgroup}c<{\egroup}@{}}

\definecolor{lightblue}{rgb}{.90,.95,1}
\definecolor{mypink1}{rgb}{0.858, 0.188, 0.478}
\definecolor{mypink2}{RGB}{219, 48, 122}
\definecolor{mypink3}{cmyk}{0, 0.7808, 0.4429, 0.1412}
\definecolor{mygray}{gray}{0.90}
\definecolor{codehighlight}{rgb}{0.95,0.8,0.8}
\definecolor{codebackground}{rgb}{0.95,0.95,0.95}
\definecolor{lightyellow}{RGB}{255,255,204}

\newtoggle{includeAcks}
\toggletrue{includeAcks}

\newtoggle{techreport}
\togglefalse{techreport}

\newtoggle{acmFormat}
\togglefalse{acmFormat}

\iftoggle{acmFormat}{
\settopmatter{printacmref=false, printccs=false, printfolios=false}
\setcopyright{none}
\pagestyle{plain} 
\renewcommand\footnotetextcopyrightpermission[1]{} 

\acmConference[]{}{}{}
}{}

\begin{document}


\title{SLAMBooster: An Application-aware Online Controller for Approximation in Dense SLAM}

\iftoggle{acmFormat}{
\iftoggle{techreport}{
  \author{Yan Pei}
  \affiliation{
    \institution{University of Texas at Austin, USA}  
  }
  \email{ypei@cs.utexas.edu}

  \author{Swarnendu Biswas}
  \authornote{This work was carried out when the author was affiliated to the University of Texas at Austin. }
  \affiliation{
    \institution{Indian Institute of Technology Kanpur, India}
  }
  \email{swarnendu@cse.iitk.ac.in}

  \author{Donald S. Fussell}
  \affiliation{
    \institution{University of Texas at Austin, USA}  
  }
  \email{fussell@cs.utexas.edu}

  \author{Keshav Pingali}
  \affiliation{
    \institution{University of Texas at Austin, USA}  
  }
  \email{pingali@cs.utexas.edu}


}{}
} 
{ 

\author{\IEEEauthorblockN{Yan Pei}
\IEEEauthorblockA{
\textit{University of Texas at Austin, USA}\\
ypei@cs.utexas.edu}
\\
\IEEEauthorblockN{Donald S. Fussell}
\IEEEauthorblockA{
\textit{University of Texas at Austin, USA}\\
fussell@cs.utexas.edu}\\
\and
\IEEEauthorblockN{Swarnendu Biswas}
\IEEEauthorblockA{
\textit{Indian Institute of Technology Kanpur, India}\\
swarnendu@cse.iitk.ac.in}
\\
\IEEEauthorblockN{Keshav Pingali}
\IEEEauthorblockA{
\textit{University of Texas at Austin, USA}\\
pingali@cs.utexas.edu}
}

\date{}
\maketitle





\begin{abstract}

\emph{Simultaneous Localization and Mapping} (SLAM) is the problem of
constructing a map of a mobile agent's environment while localizing the
agent within the map. Dense SLAM algorithms perform reconstruction and localization
at pixel granularity. These algorithms require a lot of computational power,
which has hindered their use on low-power resource-constrained devices.

Approximate computing can be used to speed up
SLAM implementations as long as the approximations do not prevent the
agent from navigating correctly through the environment. Previous studies of
approximation in SLAM have assumed that the entire trajectory of the agent
is known before the agent starts, and they have focused on \emph{offline}
controllers that set approximation knobs at
the start of the trajectory. In practice, the trajectory is usually not
known ahead of time, and allowing knob settings to change \emph{dynamically}
opens up more opportunities for reducing computation time and energy.

In this paper, we describe \emph{\name}, an \emph{application-aware}, \emph{online}
control system for dense SLAM that adaptively controls approximation knobs
during the motion of the agent. \name is based on a control technique
called \emph{proportional-integral-derivative} (PID) controller but our experiments
showed this application-agnostic controller led to an unacceptable reduction in localization accuracy. To address this problem, \name also exploits domain knowledge for controlling approximation by performing smooth surface detection and pose correction.

We implemented \name in the open-source SLAMBench framework and evaluated it on
more than a dozen trajectories from both the literature and our own study. Our
experiments show that on the average, \name reduces the computation time by 72\% and 
energy consumption by 35\% on an embedded platform, while maintaining the accuracy of localization within reasonable bounds. These improvements make it feasible to deploy SLAM on a wider range of devices.
\end{abstract}


} 

\thispagestyle{empty}

\section{Introduction}
\label{sec:intro}

Approximate computing has been shown to be useful in reducing power and energy requirements in computational science applications~\cite{loop-perforation-fse-2011,paraprox-2014, carbin-pldi-2012, rely-oopsla-2013, flexjava, enerj-pldi-2011, sampson-pldi-2014}.

Emerging problem domains like autonomous vehicles, robot navigation, augmented reality and the Internet of Things have opened up new opportunities for the use of approximate computing. Many of these applications need to run on embedded and low-power devices, so reducing their power and energy requirements permits them to be deployed on a wider range of devices for longer durations. However, they are usually streaming applications in which inputs are not provided at the start of the program as they are in computational science applications but are supplied to the application over a period of time. Approximation in such programs must be performed in an adaptive, time-dependent manner to exploit temporal properties of the streaming input.

This paper is a case study of the use of principled approximation in
Simultaneous Localization and Mapping (SLAM)~\cite{kinectfusion-uist-2011,
kinectfusion-ismar-2011, kintinuous, orbslam, ekf-slam, elasticfusion,
engel2014lsd, tateno2017cnn, kahler2015very, leutenegger2015keyframe,
fpga-soc-slam, gpu-slam, davison2007monoslam, klein2007parallel,
forster2014svo}, which is an important problem in domains
such as robot navigation, augmented reality, and control of
drones, robots and other autonomous agents. Unmanned agents have sensors like cameras or LIDAR to probe their environments. The SLAM problem is to use this sensory input to (i)
construct a map of the agent's environment (\emph{mapping}), and (ii) determine the agent's position and orientation in this environment (\emph{localization}). The mapping and localization steps are performed repeatedly as the agent explores the environment.


Dense SLAM algorithms~\cite{elasticfusion, kinectfusion-ismar-2011} require a lot of computation for several reasons. They utilize all frame pixels for reconstruction (in contrast, sparse SLAM algorithms utilize only a subset of features~\cite{orbslam}) so
there is a lot of data to process. The mapping phase requires repeated application of floating-point-intensive kernels such as stencil computations and filters~\cite{ekf-slam, kinectfusion-ismar-2011, sim2005vision} for reducing noise in incoming frames when operating in real-world environments~\cite{bodin-2016, wapco-2017}. Computationally intensive algorithms such as iterative closest point~\cite{kinectfusion-ismar-2011} and Gauss-Newton minimization~\cite{engel2014lsd} are used in the localization process. Therefore to achieve real-time performance,
dense SLAM needs a lot of computational power, and deployment of dense SLAM on battery-operated, low-power devices 
often leads to poor performance even though these devices are natural targets for SLAM.

\subsection{Approximating SLAM}

Approximate computing can be used to reduce the time and energy requirements of SLAM implementations as long as the approximations do not prevent the agent from navigating correctly through the environment. Implementations of SLAM algorithms usually expose a number of algorithmic parameters, also called \emph{knobs}, that trade off computation for accuracy of localization and mapping. Several strategies have been explored in the literature to control knobs in SLAM and other applications with streaming inputs.

{\em Offline control:} Prior work by Bodin {\em et al.} in PACT'2016 has studied
approximation in SLAM under the assumption that the entire trajectory is known
before the agent starts to move. Design space exploration for the given trajectory
is performed by executing actual trials and the results are used to select good knob settings that are used for the entire trajectory~\cite{bodin-2016}.
This is an example of {\em offline control} since knob settings are
determined once and for all before the computation begins. Subsequent studies
along this line highlighted opportunities for exploiting approximation
in SLAM algorithms~\cite{slambench-2015,wapco-2017,slam-dse}.
In most applications of SLAM however, trajectories are not known ahead of time.
Permitting knobs to be controlled adaptively during the navigation of the agent also opens up more opportunities for reducing time and energy requirements.

{\em Application-agnostic control:} Several application-agnostic control systems have been proposed for trading off power or energy consumption for performance and program accuracy~\cite{jouleguard-sosp-2015,meantime-atc-2016,poet-2015,khudia2015rumba}.
These controllers are based on general control-theoretic principles. We
investigated several such controllers but found that they do not work well for
SLAM (Sections~\ref{sec:heuristics} and \ref{sec:eval:other_comparison}).

\emph{\name}: In this paper, we present \name, an \emph{application-aware} \emph{online} control system for a popular dense SLAM algorithm called
\emph{KinectFusion}~\cite{kinectfusion-ismar-2011, kinectfusion-uist-2011}, which
allows real-time reconstruction of a room-size
environment on desktop computers. Recent innovations in dense SLAM algorithms have been implemented in \kfusion~\cite{whelan2015real, kintinuous, kahler2015very, elasticfusion}.
The contributions of this paper are as follows.
\begin{itemize}
    \item To the best of our knowledge, \name is the first controller that can
    perform online control of approximation in a SLAM algorithm.

    \item The correctness of SLAM algorithms is defined using properties of the entire trajectory. For online control, these correctness criteria must be described using quantities that can be measured during motion. Section~\ref{sec:design} shows how we accomplish this for SLAM.

    \item \name is based on the application-agnostic proportional-integral-derivative (PID) approach to control, but the PID controller is not effective in controlling SLAM by itself. We show that augmenting this controller with domain knowledge is effective in solving the SLAM control problem (Section~\ref{sec:heuristics}).


    \item We implemented \name in the open-source SLAMBench framework and evaluated it on  more than a dozen trajectories
    (Section~\ref{sec:results}). Our experiments show that on the average,
    \name reduces the computation time by 72\% and the energy consumption by 35\%
    on an embedded platform while maintaining the accuracy of the localization within bounds.
\end{itemize}

\name therefore is an important step towards enabling efficient execution of dense SLAM algorithms on a wider range of platforms; more generally, it provides a case study of how to control approximation online in streaming applications in a principled way.

\section{KinectFusion and SLAMBench}
\label{sec:background}
\label{sec:kfusion_bg}

This section describes the \kinectfusion algorithm (Section~\ref{sec:kf})
and the SLAMBench infrastructure (Section~\ref{sec:slambench_bg})
in enough detail to explain
what knobs are available and what they do. We also highlight the performance challenges in using \kinectfusion (and SLAM algorithms in general) on resource-constrained platforms
 (Section~\ref{sec:ori_perf}).

\begin{figure}[t]
  \centering
  \subfloat[Depth frame captured by an agent]{\includegraphics[scale=0.23]{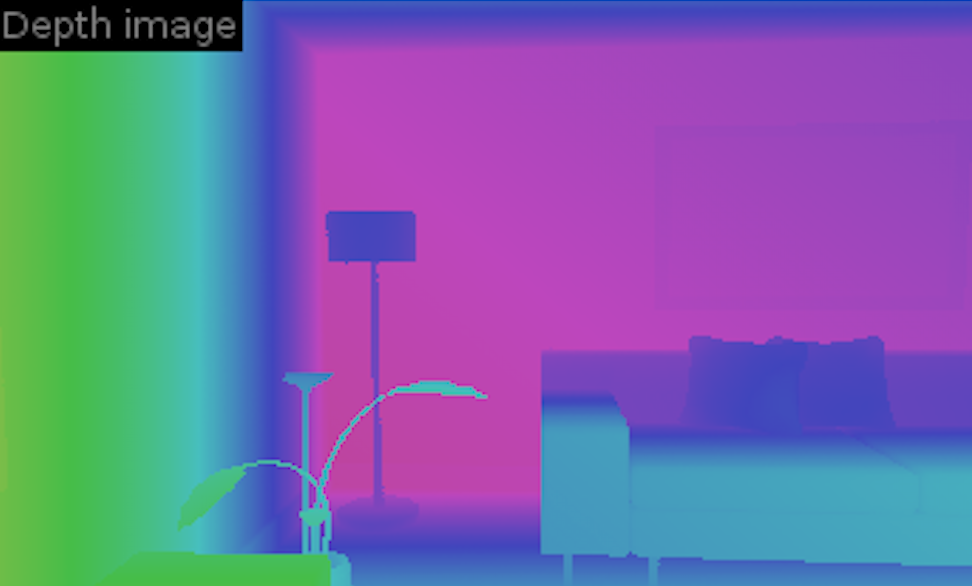}\label{fig:depth_frame_example}}
  \hspace*{5pt}
  \subfloat[3D reconstructed surface]{\includegraphics[scale=0.113]{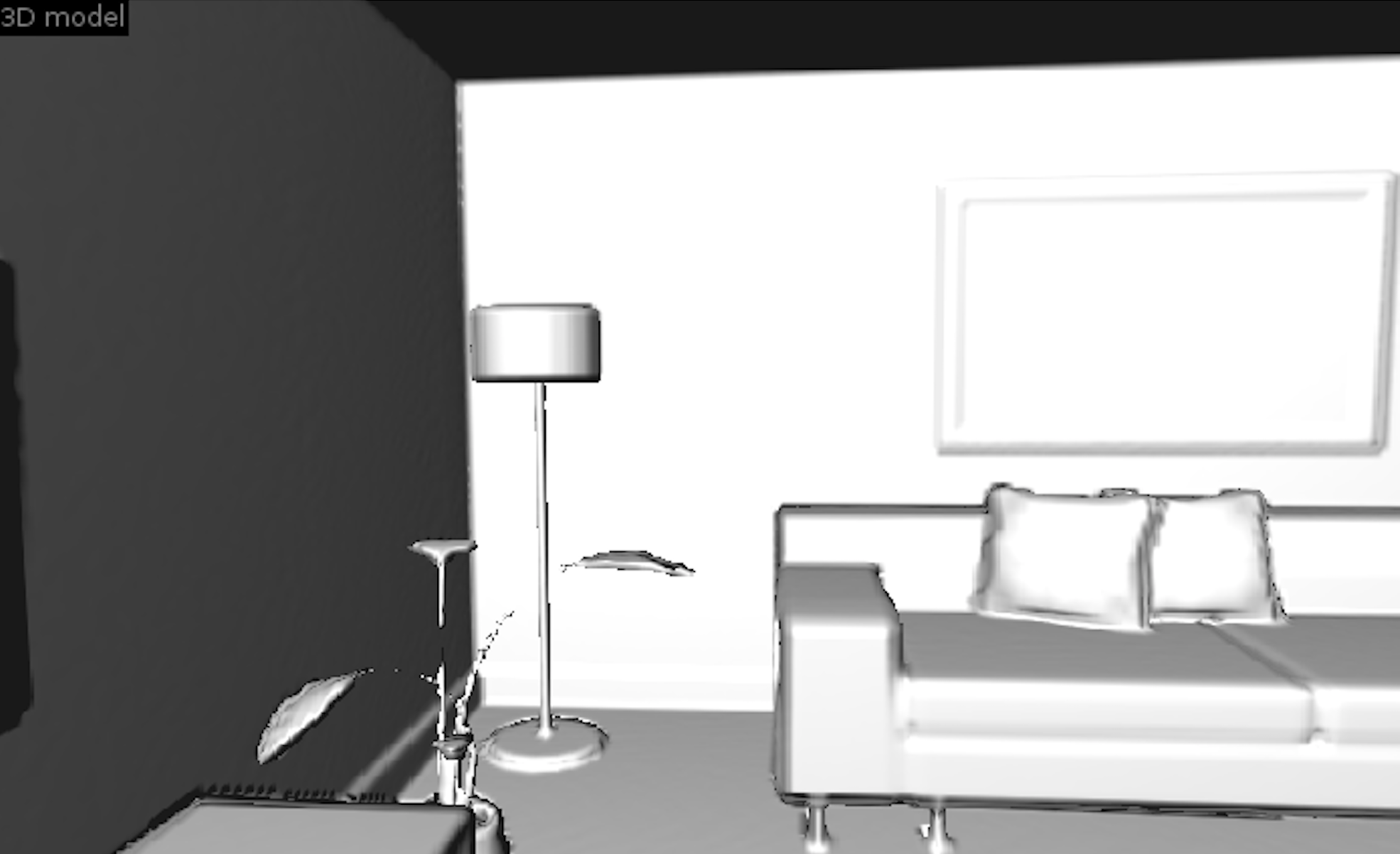}\label{fig:3D_reconstruction_example}}
  \caption{KinectFusion takes depth frames as inputs and produces a 3D reconstructed surface with the trajectory of the agent in that environment.}
  \label{fig:real_to_result_moti}
\end{figure}

\subsection{Brief Description of KinectFusion}
\label{sec:kf}

The KinectFusion algorithm performs the following high-level steps for each input frame~\cite{kinectfusion-ismar-2011,kinectfusion-uist-2011,bodin-2016}.

\begin{enumerate}
  \item \emph{Acquisition}: an input depth frame such as the one shown in Figure~\ref{fig:depth_frame_example} is read in either from a camera or from disk.

  \item \emph{Preprocessing}: depth values in the frame are normalized and a bilateral filter is applied for noise reduction.

  \item \emph{Localization}: a new estimate of the position and
        orientation (together called a \emph{pose}) of the camera is
        computed using the Iterative Closest Point (ICP) algorithm.
        The algorithm determines the difference in the alignment of the current normalized
        depth frame with the depth frame computed from the previous camera
        pose (see raycasting below).  This phase is also called
        \emph{tracking} in the SLAM 
        literature.

  \item \emph{Integration}: the existing 3D map
        is updated to incorporate the aligned data for the current
        frame using the pose determined in the tracking phase.

  \item \emph{Raycasting}: a depth frame from the new camera pose is computed
        from the global 3D map by raytracing.

  \item \emph{Rendering}: a visualization of the 3D surface is generated, as shown in Figure~\ref{fig:3D_reconstruction_example}.
\end{enumerate}




\subsection{SLAMBench}
\label{sec:slambench_bg}

The SLAMBench~\cite{slambench-2015} implementation\footnote{\url{https://github.com/pamela-project/slambench}} of the \kfusion algorithm exposes the following algorithmic parameters (knobs), which are set to certain default values in SLAMBench.

\begin{enumerate}
  \item \emph{Compute size ratio} (\knob{csr}): resolution of the depth frame used as input.

  \item \emph{Tracking rate} (\knob{tr}): rate at which tracking and localization are performed.

  \item \emph{Integration rate} (\knob{ir}): rate at which new frames are integrated to the scene.

  \item \emph{ICP threshold} (\knob{icp}): threshold for the ICP algorithm.

  \item \emph{Pyramid level iterations} (\knob{pd}): maximum number of
      iterations that ICP algorithm can perform on each level of the image pyramid.

  \item \emph{Volume resolution} (\knob{vr}): resolution at which the scene is reconstructed.

  \item \emph{$\mu$ distance} (\knob{mu}): the truncation distance in the output volume representation.
\end{enumerate}

These knobs can be tuned to optimize the computation time or energy required for
processing each frame\footnote{In our study, we ignore the image acquisition time and
rendering time since acquisition time is platform-independent and rendering is
not a necessary step in \kfusion.}. However, this tuning needs to be done under the constraint that the
output quality is acceptable since \kfusion can construct inaccurate 3D
maps or trajectories when too much approximation is introduced.


The constraints on output quality are defined as follows. The difference between the actual and the computed location of the agent at any frame is defined as \emph{instantaneous trajectory error} (ITE). One widely used quality metric used for SLAM is the {\em average trajectory error} (ATE), which is the average of the ITE over all the frames of the trajectory. Since ATE is a property of the entire trajectory, it is not known until the end of the trajectory. Furthermore, it requires knowing ground truth (\ie, the actual trajectory taken by the agent), which is usually not available except in simulated SLAM environments. \emph{Online control of knobs in SLAM therefore requires a proxy for the ATE that can be computed at each frame rather than at the end of the computation.} Section~\ref{sec:err_proxy} describes the proxy used in \name.

\subsection{Performance of \kfusion in SLAMBench}
\label{sec:ori_perf}

\begin{figure}
  \centering
  \subfloat[Computation time]{\includegraphics[scale=0.43]{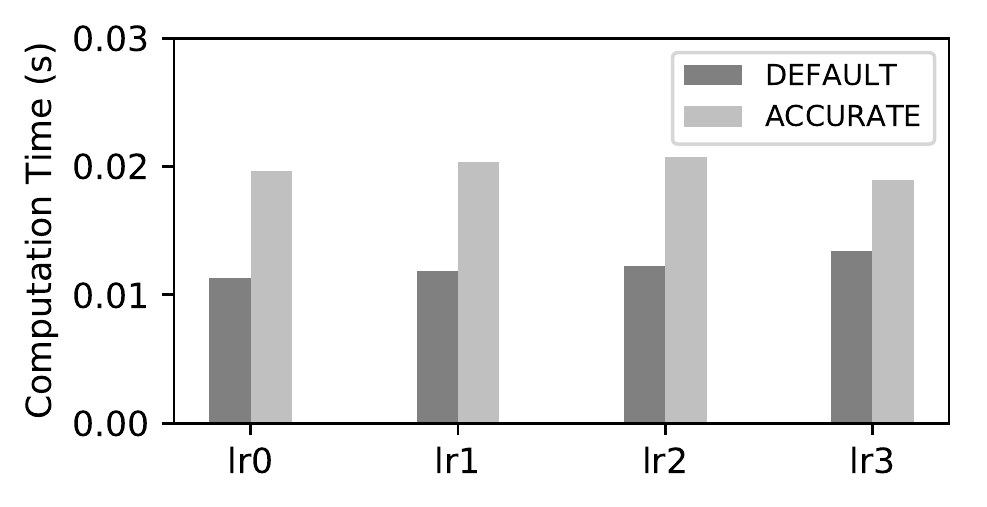}\label{fig:momentum-comptime}}
  \hspace*{5pt}
  \subfloat[ATE]{\includegraphics[scale=0.43]{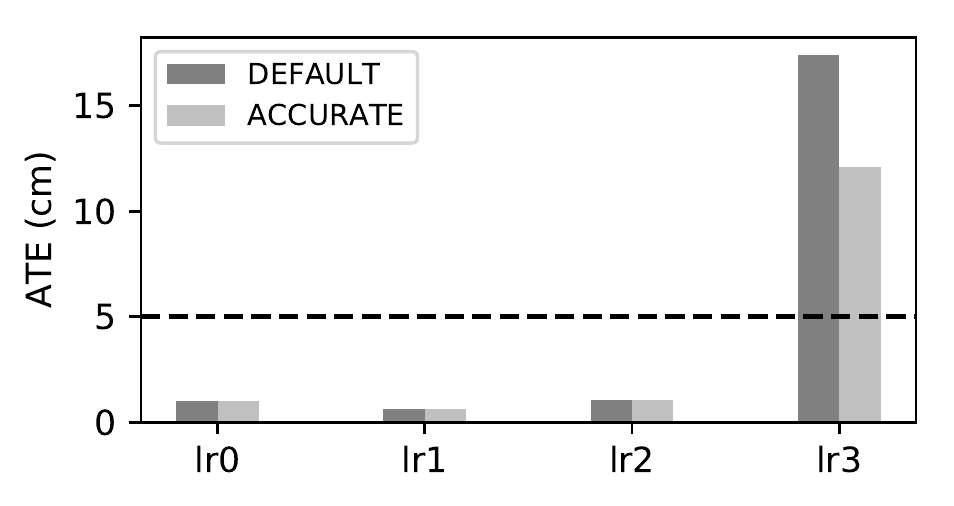}\label{fig:momentum-ate}} \\
  \vspace{-8pt}
  \subfloat[Computation time (in seconds)]{\includegraphics[scale=0.43]{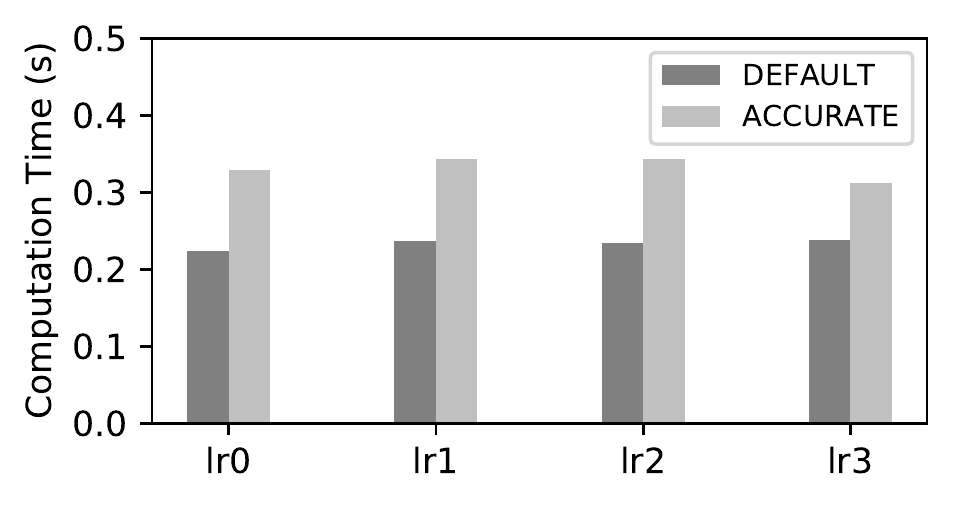}\label{fig:odroid-comptime}}
  \hspace*{5pt}
  \subfloat[ATE]{\includegraphics[scale=0.43]{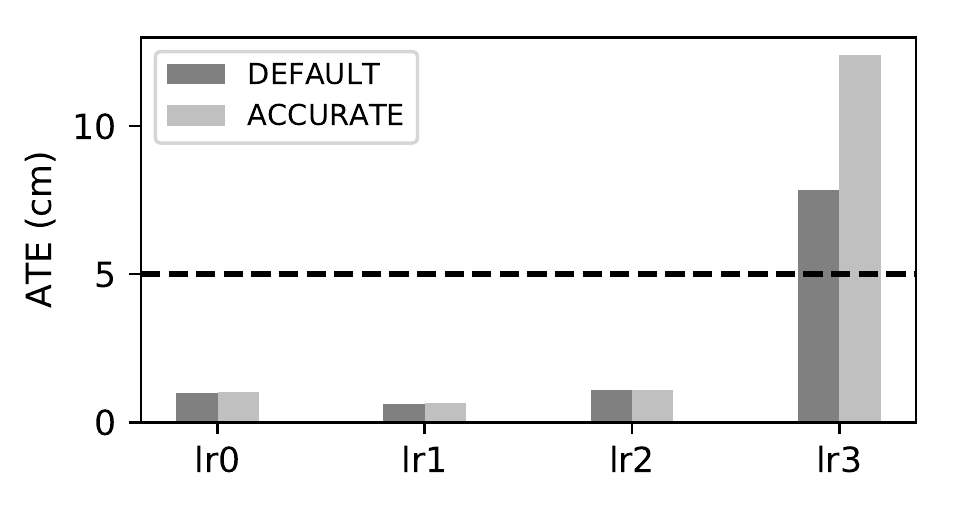}\label{fig:odroid-ate}}
  \caption{Performance of the default and the most accurate configurations of KinectFusion. The top row shows performance on an Intel Xeon system, while the bottom row shows performance with an ODROID XU4 system.}
  \vspace{-10pt}
  \label{fig:motivation}
\end{figure}

To get a sense of the performance of \kfusion with the default knob settings used
in SLAMBench, we ran an unmodified OpenCL implementation of \kfusion on two platforms with different compute capabilities. 
The first platform is an Intel Xeon E5-2630 desktop system with a Nvidia Quadro M4000 GPU, and the other is an ODROID XU4 board which is an widely used platform for emulating embedded systems. The ODROID XU4 has an octa-core Exynos 5422 big.LITTLE processor and a Mali-T628 MP6 GPU. The top row in
Figure~\ref{fig:motivation} shows the computation time per frame and ATE on the Xeon system,
while the bottom row shows these values for the ODROID system.
In each figure, the horizontal axis shows four living room trajectories named \bench{lr0}-\bench{lr3}
from the ICL-NUIM dataset~\cite{iclnuim}. The dataset includes ground truth, so it is possible to compute the ATE for these trajectories.

For computation time performance, the vertical axis shows the computation time
required by \kfusion to process each input frame. For the ATE, the vertical axis
shows the average deviation between the reconstructed trajectories and the ground
truth. In SLAM literature, an ATE of 5~cm or less is considered reasonable~\cite{bodin-2016}; the horizontal dashed lines in Figures~\ref{fig:momentum-ate} and \ref{fig:odroid-ate} show the 5~cm constraint on the ATE. The left bar, \emph{DEFAULT}, represents the default settings of the
parameters in \kfusion as set in SLAMBench.
\emph{ACCURATE} represents the most accurate configuration of the parameters for \kfusion and
thus incurs more overhead than \emph{DEFAULT}. (Note that neither \emph{DEFAULT} nor
\emph{ACCURATE} can process \bench{lr3} well because \bench{lr3} has frames
that are too difficult to be accurately tracked, so the difference in the amount of error for
\bench{lr3} introduced by \emph{DEFAULT} and \emph{ACCURATE} can be ignored.)

From Figure~\ref{fig:motivation}, we see that the best frame rate attainable by
\kfusion on the high-end Xeon system is approximately 90~fps, but it is \emph{only 3-4~fps} on the big.LITTLE embedded system. Thus, while \kfusion can achieve
real-time processing rates on high-end hardware, it performs poorly on an embedded system with constraints on resources such as hardware capabilities, energy or power consumption, and peak frequency. One way around this problem is to use approximation but this needs to be done without reducing the quality of the output to an unacceptable level. The rest of the paper explores
how this is done in \name.

\section{Design Choices in Approximation Controller}
\label{sec:design}

In this section, we describe the main choices made in the design of \emph{\name}.

\subsection{Offline vs. Online Control of Knobs}
\label{sec:off_vs_on}

An offline control system for SLAM would use fixed knob settings for the entire computation, and it would choose the knob settings using cost and error models built from training data, and features of the input trajectory. Offline control has been used successfully to control approximation in long-running compute-intensive programs~\cite{capri-asplos-2016} but there are some obvious drawbacks in using this approach for SLAM. In most applications of SLAM such as
robot motion, the environment is discovered while moving through it so the trajectory is not known before the SLAM computation begins. Furthermore, online control permits knob settings to be set adaptively, utilizing information from each frame, and this can be more efficient than setting the knobs once and for all at the start of the SLAM computation. For example, if the scene has
objects like chairs or tables, localization and integration are relatively easy
and the SLAM computation can be performed with lower precision. Conversely, when
the scene has only smooth surfaces like walls, it complicates the process of
tracking and aligning frames, and a more precise computation may be needed to
avoid large tracking error. A quantitative comparison between online control and offline
control is given in Section~\ref{sec:eval:odroid_perf}.
For these reasons, \name uses online control of approximation.


%

\subsection{Proxy for Instantaneous Trajectory Error}
\label{sec:err_proxy}


An online control system needs online metrics to monitor the performance of the application during execution. As discussed in Section~\ref{sec:background}, the usual error metric used in SLAM is the Average Trajectory Error (ATE) but this is a property of the entire trajectory so it cannot be used directly as an online error estimator. If ground truth is available, we can use the instantaneous trajectory error (ITE) but in most applications of SLAM, ground truth is not available.

To devise a proxy for ITE, we exploit the basic assumption in the \kfusion algorithm that the movement of the agent between successive frames is small (the algorithm
for localization is based on this assumption). To make use of this observation, we evaluated several plausible metrics including the \emph{inter-frame difference in depth values} and the \emph{alignment error} as online proxies for error; intuitively, large values of these metrics suggest that the scene is changing suddenly, requiring higher precision computation to avoid large tracking error. However, our experiments showed that there is no strong correlation between these metrics and the ITE. The metric which correlated best with the ITE is the \emph{velocity} of the agent. The localization phase in the \kfusion algorithm estimates the pose of the agent, and by looking at the difference in poses
between successive frames (assuming every frame is tracked), we can estimate
the velocity of the agent. We used Lasso~\cite{lasso} to confirm a positive correlation
between the velocity and the ITE for the four ICL-NUIM trajectories for which
we have the ground truth.
Therefore, we use the estimated velocity
of the agent as a proxy for the ITE.

\begin{figure}[tbph]
    \centering
    \subfloat[Ranking knobs for computation time]{\includegraphics[scale=0.5]{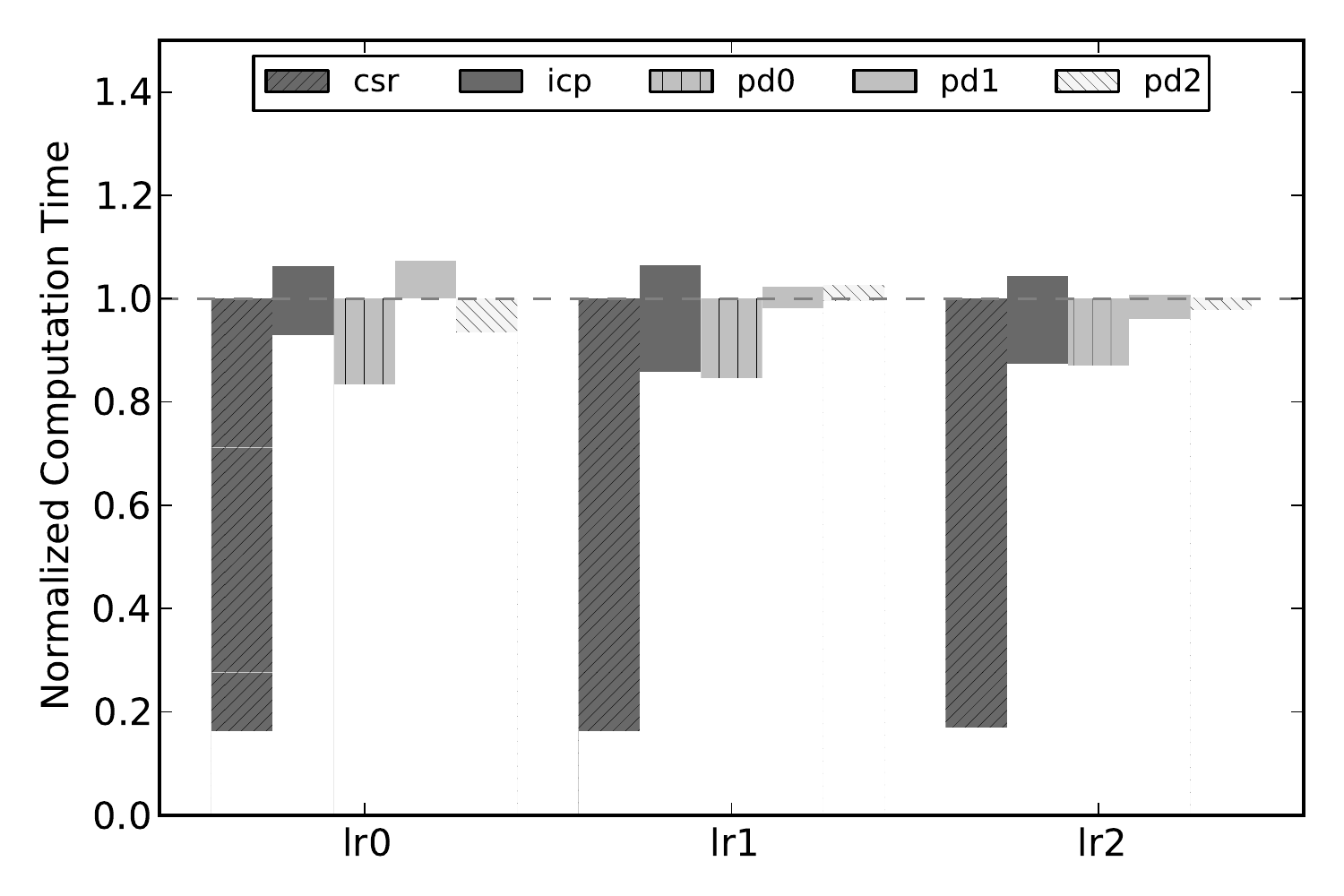}\label{fig:time_knob_rank1}}\\
    \vspace{-5pt}
    \subfloat[Ranking knobs for ATE]{\includegraphics[scale=0.5]{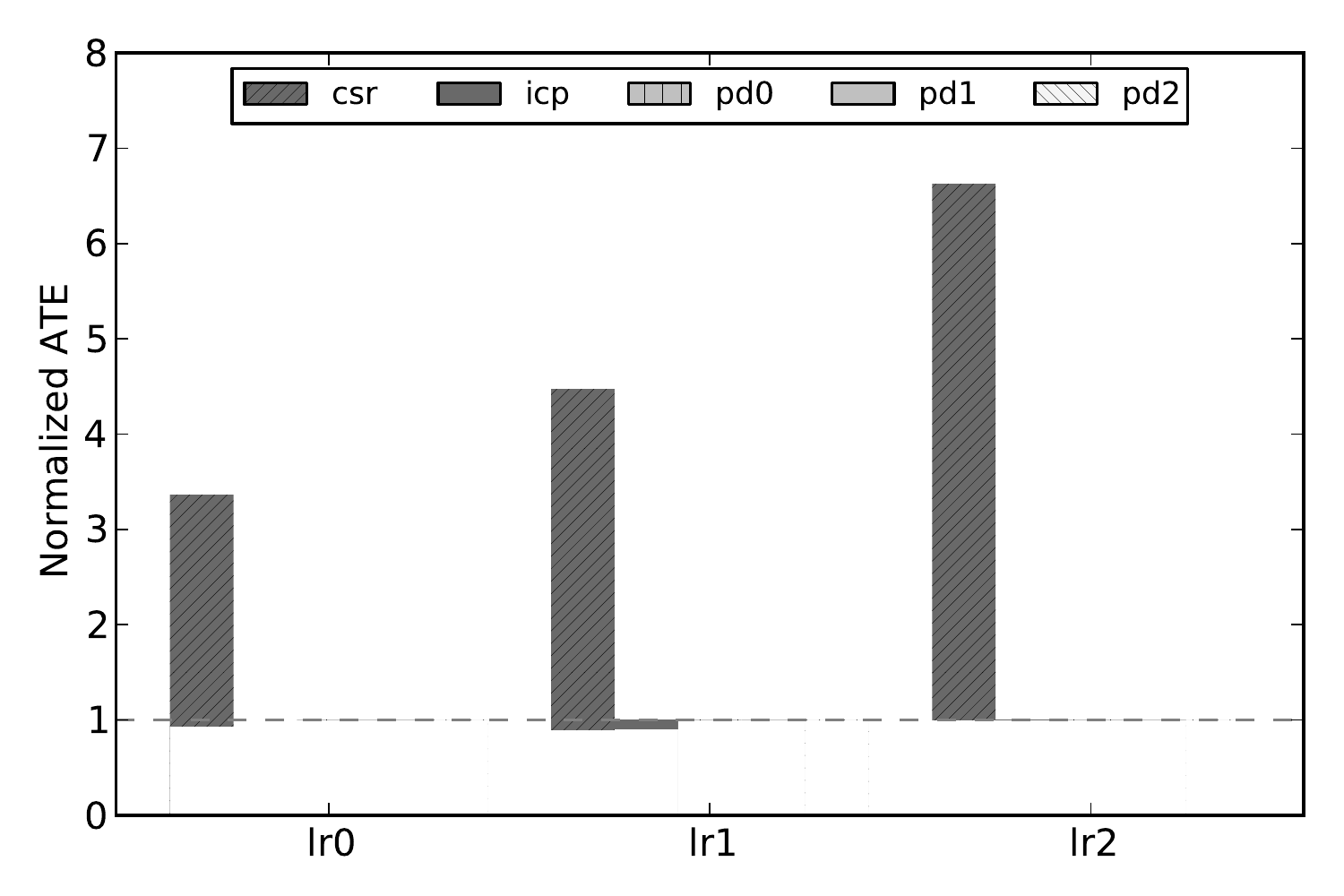}\label{fig:err_knob_rank1}}
    \caption{Ranking knobs by importance for ATE and computation time}
    \label{fig:knob_rank1}
\end{figure}

\subsection{Reducing the Knob Space}
\label{sec:knob_rank}


Precise modeling of the relationship between knob settings and computation time or error requires extensive exploration of the knob space and is input-sensitive,
which makes it intractable~\cite{bodin-2016}.
To make the control problem more tractable, we ignore knobs that are ill-suited
for online tuning, such as \knob{vr} and \knob{mu}, since they're input
dependent and require recomputing the global map data structure, which is expensive. We also do not control the tracking rate \knob{tr} or integration rate \knob{ir} and we set them to one, since every frame should be
tracked and integrated in 
an effort to not violate the assumption that the movement between successive frames is small.

We ranked the remaining knobs by their influence on ATE and computation time. These knobs are \knob{csr}, \knob{icp}, and \knob{pd}. The knob \knob{pd} has three components, referred to as \knob{pd0}, \knob{pd1}, and \knob{pd2}.
Figures~\ref{fig:time_knob_rank1} and \ref{fig:err_knob_rank1} show how computation
time and ATE change for the first three living room trajectories from the ICL-NUIM
dataset when knobs are changed
one at a time, keeping all other knobs fixed at their default values.
We find that knobs \knob{csr}, \knob{icp} and \knob{pd0} have the most impact on performance, and this finding is consistent with prior work~\cite{bodin-2016}.
Among the three, \knob{csr} has dominant impact on computation time and also significantly impacts ATE, which is intuitive since it controls the resolution of the depth image to be used
for computation. Knob \knob{pd1} and \knob{pd2} did not significantly influence
ATE or computation time, and hence are less interesting
for control. Therefore, we only use \knob{csr}, \knob{icp} and \knob{pd0}
for approximation control in \name.

\section{SLAMBooster Online Control System}
\label{sec:heuristics}



This section describes \name in stages. Section~\ref{sec:pid_ctrl} presents a
proportional-integral-derivative (PID) controller
that controls the knobs identified in Section~\ref{sec:knob_rank}.
This PID controller is successful in reducing computation time and power
consumption but it violates the error constraint by a significant
margin for some trajectories. Therefore, we improve it by exploiting
domain-specific knowledge of the \kinectfusion algorithm, using smooth surface
detection (Section~\ref{sec:wall_detection}) and pose correction (Section~\ref{sec:extrapolation}).
The computation time performance is further improved by using reduced-precision
floating-point operations in some of the computation phases (Section~\ref{sec:half_precision}).

\begin{small}
\begin{algorithm}[t]
	\caption{Online controller in \name for the \kfusion algorithm applied on each input frame Frame$_t$.}
	\label{alg:heuristic}
	\begin{algorithmic}[1]
		\newcommand\letequal{\ensuremath{\gets}\xspace}

		\State Frame$_t$ \letequal \texttt{KF.acquisition}()\label{alg:acquisition}
		\If {$t$ $\leq$ {\small BOOTSTRAP\_FRAMES}}
		\Comment{Do not approximate}
		\State Knob$_t$ \letequal Knob$_{t-1}$\label{alg:start}
		\State Frame$_t$ \letequal \texttt{KF.preprocessing}(Frame$_t$, Knob$_t$)
		\State Pose$_t$ \letequal \texttt{KF.tracking}(Frame$_t$, Knob$_t$)
		\State \texttt{KF.integration}(Pose$_t$, Frame$_t$, Knob$_t$)
		\State \texttt{KF.raycasting}(Pose$_t$, Knob$_t$)
		\State \texttt{KF.rendering}(Pose$_t$, Knob$_t$)
		\label{alg:end}
		\Else \Comment{Check for approximation opportunities}

        \State
        \State {Knob$_t$ \letequal PID (V$_{t-1}$, V$_{ref}$)}\label{alg:pid} \Comment{PID controller}
		\State \colorbox{lightblue}{surface\_trigger} \letequal\colorbox{lightblue}{\texttt{SurfaceDetection}(Frame$_t$)}\label{alg:features}

        \If{\colorbox{lightblue}{surface\_trigger}\label{alg:surface_trigger} \textbf{or} \colorbox{lightblue}{correction\_trigger$_{t-1}$}}\label{alg:correction_trigger}
        \State \colorbox{lightblue}{Knob$_t$ \letequal Knob$_{t}$ - 1} \Comment{Increase precision}
		\EndIf\label{alg:vel-end}
		\State

		\State Frame$_t$ \letequal \texttt{KF.preprocessing}(Frame$_t$, Knob$_t$)
		\State Pose$_t$ \letequal \texttt{KF.tracking}(Frame$_t$, Knob$_t$)\label{alg:pose}
		\State

		\State {V$_t$ \letequal Pose$_t$ - Pose$_{t-1}$}\Comment{Compute velocity}\label{alg:comp-vt}

        \State \colorbox{lightblue}{correction\_trigger$_t$ {\letequal} (V$_t$ {>} CORRECTION\_THRES)}
		\If{\colorbox{lightblue}{correction\_trigger}}\label{alg:exp-start}
		\State \colorbox{lightblue}{Pose$_t$ \letequal T$_{t-1}$ * Pose$_{t-1}$}
		\State \colorbox{lightblue}{T$_t$ \letequal T$_{t-1}$}
		\Else
		\State \colorbox{lightblue}{T$_t$ \letequal Pose$_t$ * Pose$_{t-1}^{-1}$}
		\EndIf\label{alg:exp-end}
		\State

		\State \texttt{KF.integration}(Pose$_t$, Frame$_t$, Knob$_t$)
		\State \texttt{KF.raycasting}(Pose$_t$, Knob$_t$)
		\State \texttt{KF.rendering}(Pose$_t$, Knob$_t$)

		\EndIf
	\end{algorithmic}
\end{algorithm}
\end{small}


\subsection{PID Controller}
\label{sec:pid_ctrl}


In its simplest form, a PID controller~\cite{nise94} is like an automobile cruise control - it adjusts knob settings by looking at the difference between a reference value and a value that is derived from the state of the system. This is a proportional (P) controller. To make control more smooth, we can take history into account by considering also the integral (I) of this difference over a time window, and we can make the controller more reactive by considering the derivative (D) of this difference~\cite{nise94}.

Algorithm~\ref{alg:heuristic} shows the pseudocode for \name and incorporates
the refinements discussed later in this section. \code{KF} in Algorithm~\ref{alg:heuristic}
stands for \kfusion, and lines showing operations on \code{KF}
represent logic from the unmodified \Kfusion algorithm. For simplicity, knobs
settings are described in discrete levels. Increasing the level means
introducing more approximation, or vice versa.
The highlighted code implements optimizations to the baseline controller, and should be ignored for now.

\begin{itemize}

	\item Line~\ref{alg:acquisition} shows an input frame \code{Frame$_t$} is
        acquired from the I/O device. The rest of Algorithm~\ref{alg:heuristic}
        shows computations performed on \code{Frame$_t$}. Frame pixels are depth values.


	\item After acquiring the frame, the controller determines whether it is
        still in the bootstrap phase. No approximation is done for the first few
        frames to allow \kfusion to initialize an accurate global 3D map, so the
        original \kfusion algorithm is executed (lines~\ref{alg:start}--\ref{alg:end}).

    \item If SLAM is not in the bootstrap phase, the controller picks the knob
        settings (denoted by Knob$_t$) based on a PID controller (line~\ref{alg:pid}),
        which compares the agent's current velocity (denoted by V$_t$) and a reference
        velocity (denoted by V$_{ref}$). V$_t$ is estimated using the difference
        between the poses in the current frame and the previous frame
        (line~\ref{alg:comp-vt}). The level of the knobs is set
        proportionately to the difference between V$_t$ and V$_{ref}$ if V$_t$
        is smaller than V$_{ref}$. A larger difference means more approximation.
        When V$_t$ is larger than V$_{ref}$, the most accurate knob setting will
        be applied because higher velocity usually implies higher ITE.
        After estimated by the proportional (P) part, the knob settings will
        be further refined by the integral (I) part and the derivative (D) part.

    \item The I part is a sliding window of the history of V$_t$ and tracks
        the average velocity over the current window. It adjusts the knob
        level by comparing the average velocity with a predefined threshold.
	    The length of the velocity window is set to the length of the bootstrap phase.  On the other hand, the D part checks whether the velocity is consistently
        increasing or  decreasing over the same sliding window that the I part
        uses. If the velocity is consistently decreasing, the knob level can
        be increased to exploit more approximation and vice versa.

\end{itemize}

Our experiments showed that although this PID controller is effective in
reducing the computation time by more than half, it fails to meet the trajectory
error constraint for many of our real-world trajectories (see Section~\ref{sec:optimization_impact}).
This application-agnostic approach itself is not enough to
balance the trade-off between computation time and localization accuracy for the
following two reasons:
\begin{itemize}
    \item This controller does not utilize any information from
        the input frame. The controller would introduce approximation too
        aggressively when V$_t$ is low, while low V$_t$ doesn't necessarily mean
        that the next frame is easy to track. For example, smooth surfaces are
        much more difficult to track than scenes with table and chairs.

    \item For several reasons such as noisy frames, SLAM algorithms may produce a wrong pose that differs substantially from the previous pose. This controller cannot react to such potential incorrect estimations.
\end{itemize}

Next, we improve the controller by incorporating domain knowledge.

\subsection{Smooth Surface Detection}
\label{sec:wall_detection}

In the tracking and integration phases, \kfusion merges information from the
current depth frame with the global 3D map. Each incoming depth frame in
\kfusion is an abstraction of the scene at which the camera is pointing. If the
scene has objects like chairs and tables, it is easier
for the algorithm to align and integrate a depth frame with the existing 3D map than if
the scene is a blank wall for example. Therefore, it is desirable to adapt the
level of approximation to the scene: the smoother the surface is, the more
the accuracy that is needed.

\begin{figure}
\centering
\includegraphics[scale=0.3]{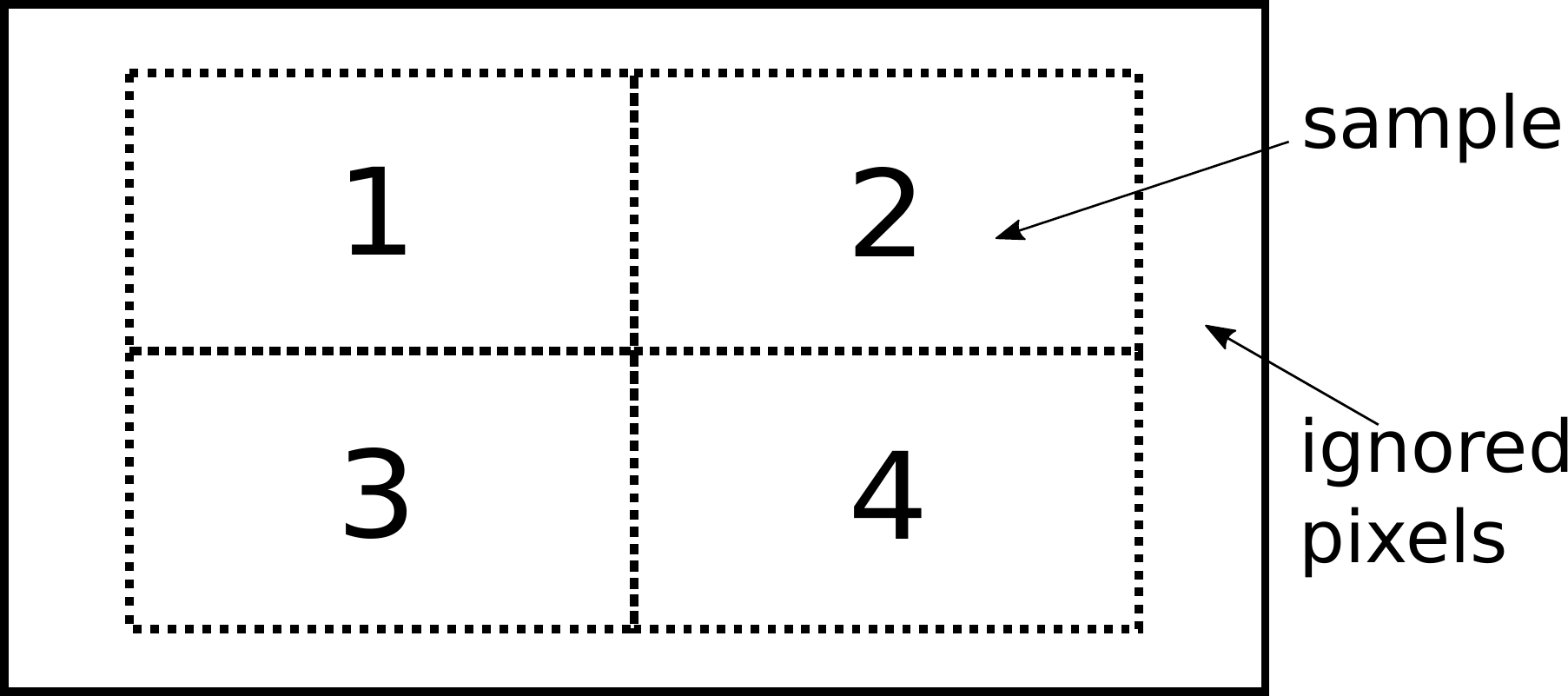}
\caption{Smooth surface detection heuristic. We compute the intra-sample
deviation from samples in the depth image.}
\label{fig:wall-detection}
\end{figure}

The degree of smoothness of a frame is represented by calculating the standard
deviations of representative regions of a frame. Lower standard deviation means
smoother surfaces.
To implement this idea, we sample the four quadrants of the input depth frame,
leaving out pixels at the margins of the frame since the Kinect sensor is known
to potentially produce invalid depth pixels at the periphery
(Figure~\ref{fig:wall-detection}). To reduce computational overhead, a
\emph{fixed} number of pixels are sampled from each quadrant, independent of the
resolution of the frame; this number is chosen so that at the lowest resolution,
all pixels are sampled. At the second lowest resolution, every other pixel is
sampled, etc. We then compute the standard deviation of depth values within each quadrant. If all these standard deviations are below some threshold value, the camera might be pointing to a smooth surface so the control system increases the precision of \kfusion computation.

Algorithm~\ref{alg:heuristic} shows the augmented control system using smooth
surface detection. The \code{SurfaceDetection} function implements smooth
surface detection (line~\ref{alg:features}, Figure~\ref{fig:wall-detection}).
This additional information is used by the controller to manipulate knobs
(line~\ref{alg:surface_trigger}). Our experiments indicate that compared with
the PID controller, this improves the ATE with \emph{little additional} computation overhead.

\subsection{Pose Correction}
\label{sec:extrapolation}

The final enhancement we make to the basic PID control system is to use
a simple form of Kalman filtering~\cite{Grewal:2014, kalman-filter-arxiv} to recompute the pose when it appears that the agent has made a sudden movement. Informally, Kalman filtering is a method for combining a number of uncorrelated estimates of some unknown quantity to obtain a more reliable estimate. In many practical problems, the unknown quantity is the state of a dynamical system, and there are two estimates of this state at each time step, one from a model of state evolution and one from measurement, that are combined using Kalman filtering.

In the context of SLAM, the unknown state is the pose of the agent. When a frame is processed by \KFusion, the tracking module uses the measured depth values in the frame to provide an estimate of the new pose, as shown in line~\ref{alg:pose} in Algorithm~\ref{alg:heuristic}. However, if this estimated pose differs substantially from the pose in the previous frame, it violates the assumption that the movement of the agent between successive frames should be small. This indicates that
\kfusion has potentially inferred an inaccurate pose, and the pose estimate from the tracking module may be unreliable.
In the spirit of Kalman filtering, we use a simple model to estimate the pose
if the estimate from the measurement produced by the tracking module is substantially different from the pose in the previous frame.
Lines~\ref{alg:exp-start}--\ref{alg:exp-end} in Algorithm~\ref{alg:heuristic} show the pseudocode. \kfusion represents the live 6DOF camera pose estimate by a rigid body \emph{transformation matrix}. T$_{t-1}$ represents the transformation matrix calculated at frame $t-1$ when Pose$_{t-1}$ is computed from Pose$_{t-2}$. The logic compares $V_t$ with a threshold to check whether correcting Pose$_t$ is required. If the velocity is below the threshold, the matrix T$_t$ will be calculated using the current and the previous pose. On the other hand, if the difference in poses is abnormally large,
Pose$_t$ is recomputed by applying T$_{t-1}$ to Pose$_{t-1}$, following
the assumption that the movement between successive frames should be small. 
Downstream \kfusion kernels work using this \emph{corrected} estimate of Pose$_t$.


Section~\ref{sec:optimization_impact} shows that correcting pose estimations in this way improves the accuracy of the trajectory reconstruction substantially, with minimal control overhead.

\subsection{Reduced-precision Floating-point Format}
\label{sec:half_precision}

Finally, we explored the benefit of using half-precision floating-point
numbers instead of the default single-precision floating-point numbers.
Half-precision format can potentially improve vector operation efficiency and
cache miss rates. OpenCL extension \code{cl\_khr\_fp16} has support for
half scalar and vector types as built-in types that can be used for arithmetic
operations, type casts, \emph{etc}.

Some phases in \kfusion such as \emph{Localization} and \emph{Integration}
have constants that are too small to be represented in half-precision format.
Therefore, we manually transformed only the \emph{Raycasting} and
\emph{Preprocessing} phases to use half-precision format. Since these phases perform
a large number of vector operations, use of reduced precision can be beneficial~\cite{wapco-2017}.







\begin{figure*}[t]
	\centerline{
		\subfloat[Absolute ATE.]
		{\includegraphics[width=1.0\linewidth]{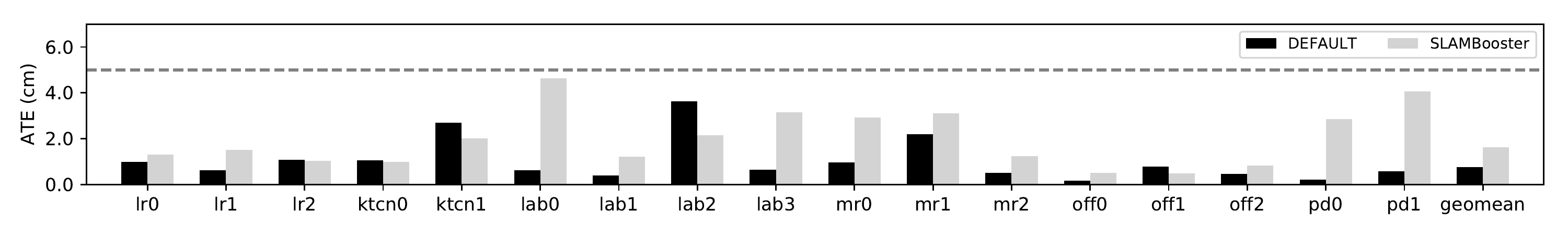}\label{fig:err_odroid}}}
	\vspace{-1pt}
	\centerline{
		\subfloat[Average Computation Time Per Frame.]
		{\includegraphics[width=1.0\linewidth]{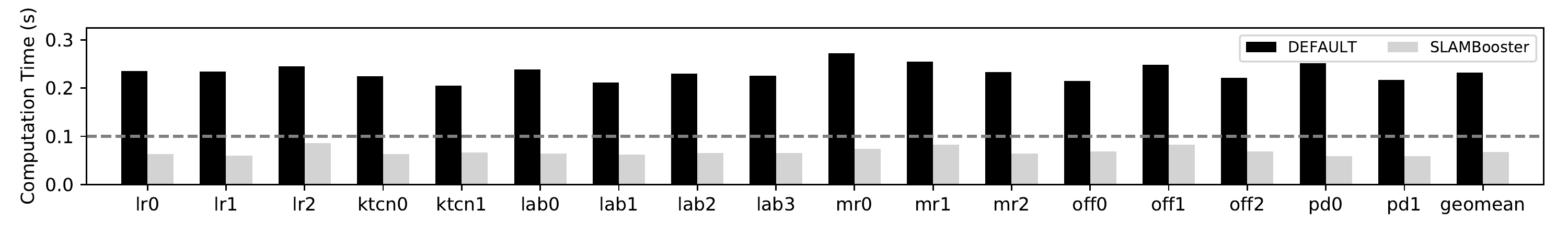}\label{fig:time_odroid}}}
	\vspace{-1pt}
	\centerline{
		\subfloat[Average Energy Consumption Per Frame.]
		{\includegraphics[width=1.0\linewidth]{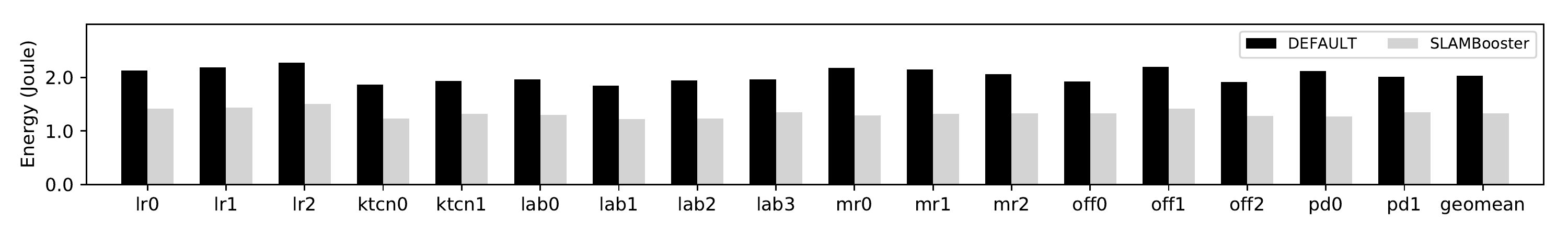}\label{fig:energy_odroid}}}
	\caption{ATE, computation time and energy performance on embedded platform}
    \vspace{-10pt}
	\label{fig:perf_odroid}
\end{figure*}

\section{Experimental Results}
\label{sec:results}

This section evaluates the benefits and effectiveness of \name
for approximating \kfusion.

\subsection{Methodology}
\label{sec:eval:methodology}


We implemented  \name in the open-source SLAMBench~\cite{slambench-2015}
infrastructure\footnote{\url{https://github.com/pamela-project/slambench}}.

\paragraph*{Platform}



Figure~\ref{fig:motivation} shows that unmodified \kinectfusion achieves good performance ($\sim$90~fps) on a high-end Intel Xeon E5-2630 system with a Nvidia Quadro M4000 GPU. Our experiments with the Intel Xeon system show that \name is able to improve the performance \emph{further} ($\sim$200~fps) without failing the accuracy constraint. We do not show results with the Intel Xeon system for lack of space. Instead, we present detailed results with \name on a low-power embedded environment using an ODROID XU4 board with a Samsung Exynos 5422 octa-core processor. The Exynos processor has four Cortex-A15 cores running at 2 GHz and four Cortex-A7 cores running at 1.4 GHz, and has 2 GB LPDDR3 RAM. The XU4 board is equipped with a Mali-T628 MP6 GPU that supports OpenCL 1.2, and runs Ubuntu 16.04.4 LTS with Linux Kernel 4.14 LTS. The XU4 board does not have on-board power monitors, we use a SmartPower2\footnote{\url{https://wiki.odroid.com/accessory/power_supply_battery/smartpower2}} device to monitor energy consumption for the whole board.

\paragraph*{Benchmark trajectories}

SLAMBench supports the ICL-NUIM RGB-D living room  dataset\footnote{\url{http://www.doc.ic.ac.uk/~ahanda/VaFRIC/iclnuim.html}},
which is used for benchmarking SLAM algorithms~\cite{iclnuim}.
This dataset is obtained by using the Kintinuous
system~\cite{kintinuous} with ground truth trajectory. Each scene has
several synthetically-generated trajectories.
We excluded benchmark \bench{lr3} from our experiments since even the \emph{ACCURATE} configuration cannot meet the error constraint (Figure~\ref{fig:motivation}).
Therefore, we use three trajectories with the living room scene, referred to as \bench{lr0},
\bench{lr1} and \bench{lr2} in this paper.

To increase the diversity of trajectories, we used a first-generation Kinect camera to collect fourteen additional trajectories from real-world scenes
in an indoor environment. The fourteen trajectories are: \bench{ktcn0},
\bench{ktcn1}, \bench{lab0}, \bench{lab1}, \bench{lab2}, \bench{lab3},
\bench{mr0}, \bench{mr1}, \bench{mr2}, \bench{off0},
\bench{off1}, \bench{off2},  \bench{pd0}, and \bench{pd1}. All
the inputs are collected at 30 fps with resolution 640x480.
We do not have ground truth for these trajectories, hence we use the trajectory
computed by the most accurate setting of \kinectfusion as a stand-in for ground
truth.
We have verified using SLAMBench GUI 
that \KFusion is able to rebuild the trajectories and 3D map correctly.
Column 2 in Table~\ref{tb:results-odroid} shows the length of each trajectory.

The extended SLAMBench implementation and collected trajectories will be made publicly available when the paper is published.

\subsection{Performance of \name}
\label{sec:eval:odroid_perf}

Figure~\ref{fig:perf_odroid} shows the ATE, average computation time per frame,
and average energy consumption per frame for the benchmark trajectories on
the embedded platform. The figures compare two configurations: using the default knob settings
in SLAMBench, and using \name to control knobs online. Each bar is the average of three trials. Performance numbers are shown in a tabular form in Table~\ref{tb:results-odroid}.

For most trajectories, ATE is higher when knobs are controlled with \name,
as expected. Nevertheless, all individual ATEs are less than 5~cm and therefore
meets the required quality constraint. Figure~\ref{fig:time_odroid} shows the
average computation time for each frame. The average speedup of \name over
\emph{DEFAULT} is 3.6x and achieves a throughput of 15~fps.
Although this frame rate does not meet real time constraint, computer graphics researchers
consider 15~fps to be reasonable for providing smooth user experience.

Figure~\ref{fig:energy_odroid} shows another important metric, the average
energy consumed for processing a frame. The idle power dissipation of an
ODROID XU4 board is about $\sim$4~Watt. The energy saving for each frame is
about $\sim$35\%. The reduction in energy per frame is not as significant as the
reduction in computation time per frame because of the frequent up and down
scaling of certain data structures, required while tuning the \knob{csr} knob.



\paragraph{Control overhead}
\label{sec:eval:ctrl_overhead}
The overhead of \name arises mainly from data structure down sampling
and up sampling when tuning knob \knob{csr}. Measurements show that the average overhead introduced by the controller
is $\sim$1~ms on ODROID XU4 board, which is $\sim$1\% of the total computation time.
Therefore, the overhead of the control logic is negligible.

\begin{table*}[t]
  \renewcommand\sfsmaller{}
  \centering
  \newcommand{\n}{\ensuremath{n}\xspace}
  \newcommand{\f}{\ensuremath{\mathit{f}}\xspace}
  \renewcommand\k[1]{#1}
  \newcommand\num[1]{0.#1}
  \newcommand\snum[1]{0.0#1}
  \begin{footnotesize}
    \begin{tabular}{l|r|Hrrr|Hrrr}
      & & \multicolumn{4}{c|}{\textbf{Default \Kfusion}} & \multicolumn{4}{c}{\textbf{\name}} \\
      & \multirow{2}{*}{\textbf{\# frames}} & \textbf{\% frames} & {\bf Comp.} & {\bf ATE} &  {\bf Energy (J)/} & \textbf{\% frames} & {\bf Comp.} &  {\bf ATE} & {\bf Energy (J)/} \\
      & & \textbf{tracked} & \textbf{time (ms)} & \textbf{(cm)} & \textbf{frame} & \textbf{tracked} & \textbf{time (ms)} & \textbf{(cm)} & \textbf{frame} \\
      \midrule
      \bench{lr0} & 1510 & 100 & 235 & 0.98 & 2.13 & 100 & 62 & 1.30 & 1.41 \\
      \bench{lr1} & 967 & 100 & 234 & 0.61 & 2.19 & 100 & 60 & 1.51 & 1.44 \\
      \bench{lr2} & 882 & 100 & 246 & 1.07 & 2.28 & 100 & 86 & 1.03 & 1.50 \\
      \midrule
      \bench{ktcn0} & 1550 & 100 & 225 & 1.05 & 1.87 & 100 & 63 & 0.97 & 1.23 \\
      \bench{ktcn1} & 800 & 100 & 204 & 2.70 & 1.94 & 100 & 66 & 2.00 & 1.31 \\
      \bench{lab0} & 800 & 100 & 239 & 0.63 & 1.96 & 100 & 64 & 4.62 & 1.30 \\
      \bench{lab1} & 1250 & 100 & 212 & 0.39 & 1.85 & 100 & 62 & 1.21 & 1.22 \\
      \bench{lab2} & 1250 & 100 & 230 & 3.63 & 1.95 & 99.5 & 65 & 2.14 & 1.23 \\
      \bench{lab3} & 1250 & 100 & 226 & 0.65 & 1.96 & 100 & 65 & 3.14 & 1.35 \\
      \bench{mr0} & 929 & 100 & 273 & 0.97 & 2.18 & 100 & 74 & 2.93 & 1.28 \\
      \bench{mr1} & 1494 & 100 & 255 & 2.19 & 2.15 & 100 & 82 & 3.10 & 1.31 \\
      \bench{mr2} & 1450 & 100 & 236 & 0.49 & 2.06 & 100 & 64 & 1.22 & 1.33 \\
      \bench{off0} & 750 & 100 & 215 & 0.17 & 1.92 & 100 & 69 & 0.52 & 1.33 \\
      \bench{off1} & 1050 & 100 & 248 & 0.78 & 2.20 & 100 & 82 & 0.49 & 1.41 \\
      \bench{off2} & 1200 & 100 & 221 & 0.45 & 1.91 & 100 & 68 & 0.83 & 1.28 \\
      \bench{pd0} & 1600 & 100 & 251 & 0.21 & 2.12 & 99.0 & 59 & 2.85 & 1.26 \\
      \bench{pd1} & 1500 & 100 & 216 & 0.57 & 2.01 & 100 & 58 & 4.06 & 1.34 \\
      \midrule
      \textbf{geomean} & & 100 & 232 & 0.75 & 2.04 & 99.9 & 67 & 1.63 & 1.32 \\
    \end{tabular}
  \end{footnotesize}
  \caption{Detailed results with input trajectories controlled by the \name on the ODROID XU4 platform.
  \vspace{-10pt}
  }
  \label{tb:results-odroid}
\end{table*}

\begin{figure*}
	\centering
	\subfloat[3D surface reconstructed by \heuristic{ACCURATE}]{\includegraphics[scale=0.41]{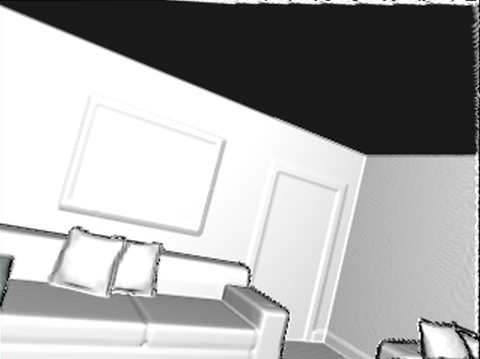}\label{fig:best_render}}
	\hspace*{8pt}
	\subfloat[3D surface reconstructed by \name]{\includegraphics[scale=0.41]{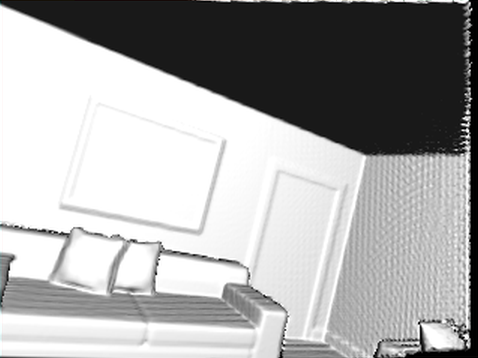}\label{fig:heuristic_render}}
	\hspace*{8pt}
	\subfloat[3D surface difference]{\includegraphics[scale=0.41]{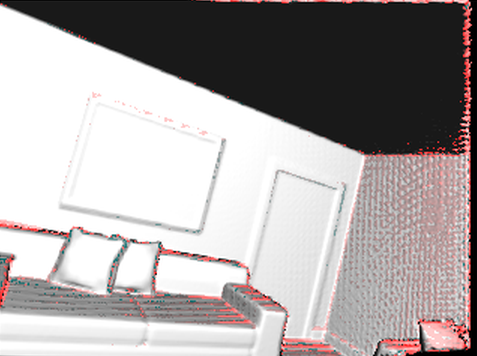}\label{fig:diff_render}}
	\caption{Evaluation of the quality of 3D maps}
    \vspace{-5pt}
	\label{fig:3d_render}
\end{figure*}

\paragraph{3D map comparison}
\label{sec:eval:3d_comparison}

Since SLAM  is used for navigation, the 3D map is mainly a mean to an end rather than an end in itself; nevertheless it is interesting to study the quality of the
3D map produced by \name.
Figure~\ref{fig:3d_render} compares a typical global 3D map built by using the most accurate knob settings (Figure~\ref{fig:best_render}) with the one built by using
\name (Figure~\ref{fig:heuristic_render}). Figure~\ref{fig:diff_render}
is a diff of these two maps in which pixels that are substantially different
are marked in red. 
We see that using \name does not impact the quality of the 3D map substantially.

\paragraph{Use only \knob{csr} in control}
\label{sec:eval:csr_control}
Since knob \knob{csr} dominates the knob space, it is interesting to see
how the controller performs without knob \knob{icp} and \knob{pd0}. Our
experiment shows that the saving in computation time and energy with only
\knob{csr} is 58\% and 29\% respectively, compared to 72\% and 35\% with
\knob{icp} and \knob{pd0}. Having more knobs in the controller helps improve the
performance of \name.

\subsection{Impact of Optimizations}
\label{sec:optimization_impact}

\begin{figure*}[t]
	\centerline{
		\subfloat[Absolute ATE.]
		{\includegraphics[width=1.0\linewidth]{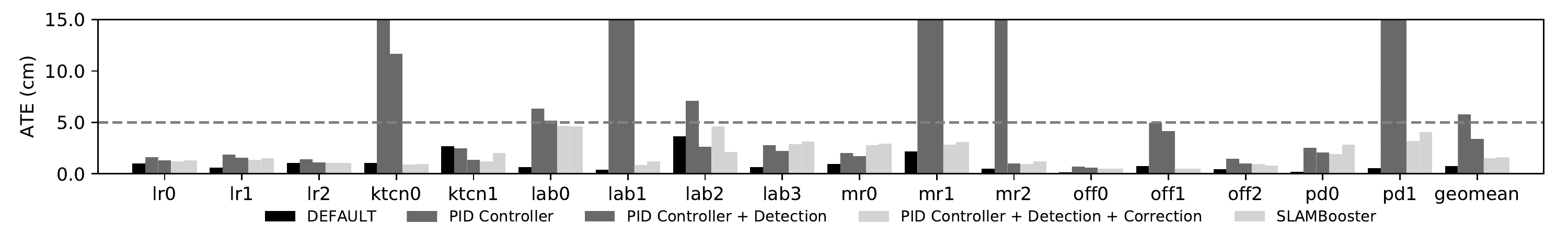}\label{fig:err_incre_opt}}}
	\vspace{-1pt}
	\centerline{
		\subfloat[Average Computation Time Per Frame.]
		{\includegraphics[width=1.0\linewidth]{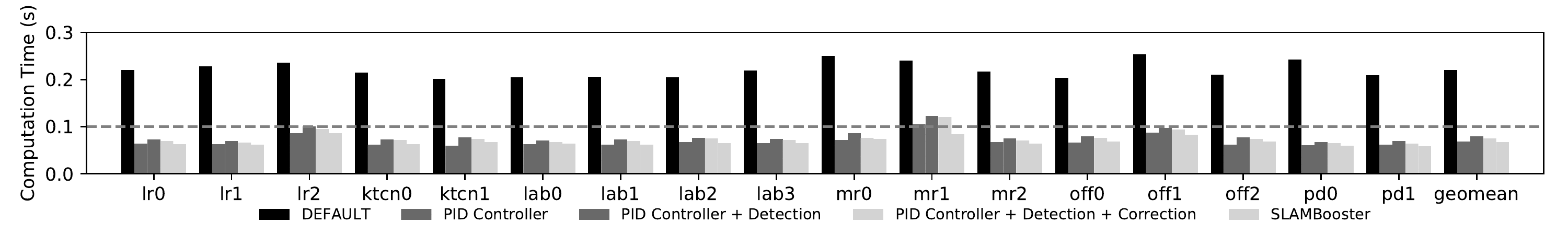}\label{fig:time_incre_opt}}}

	\caption{Incremental optimization impact on ATE and computation time}
	\vspace{-10pt}
	\label{fig:incre_impact}
\end{figure*}

\begin{figure*}[thbp]
	\centerline{
        \subfloat[Trajectory Error]
		{\includegraphics[width=1.01\linewidth]{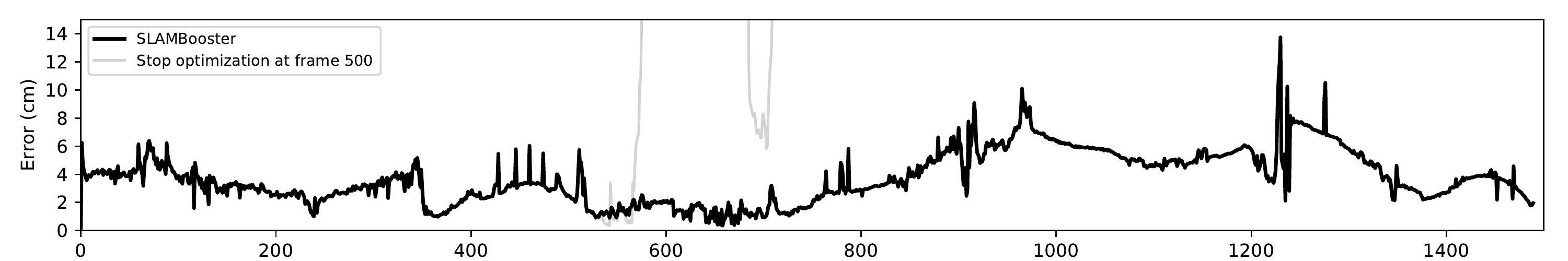}
			\label{fig:knob_activity_perf}}}
	\vspace{-2pt}
	\centerline{
		\subfloat[Knob Activity: \knob{csr}]
		{\includegraphics[width=1.0\linewidth]{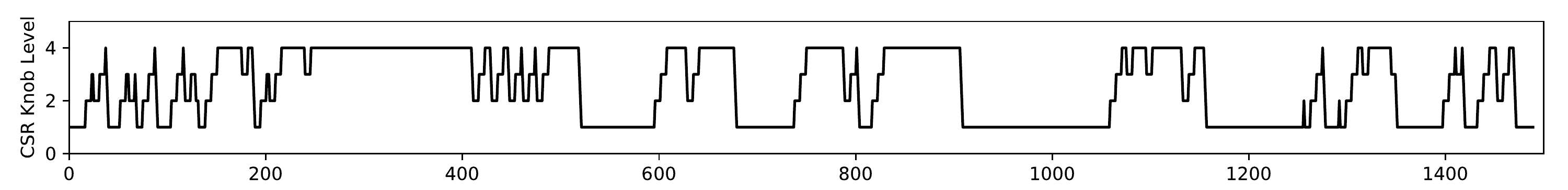}
			\label{fig:knob_activity_csr}}}
	\vspace{-2pt}
	\centerline{
		\subfloat[Trigger Activity: Surface Trigger and Pose Correction Trigger]
		{\includegraphics[width=1.0\linewidth]{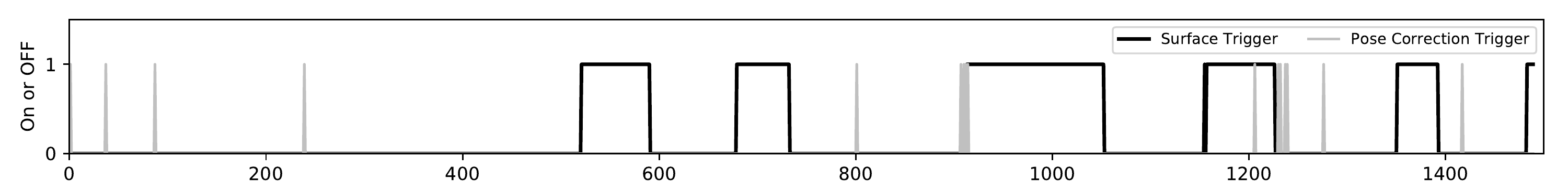}
			\label{fig:knob_activity_trigger}}}
	\caption{Trajectory error, knob activity and trigger activity in chronological order for benchmark: \bench{mr1}}
	\vspace{-10pt}
	\label{fig:knob_activity}
\end{figure*}

Figure~\ref{fig:incre_impact} shows the incremental impact of the different
optimizations on the PID controller. The controller
configurations are listed below.

\begin{itemize}
	\item \emph{DEFAULT} setting provided by SLAMBench
	\item PID control
	\item PID control + smooth surface detection
	\item PID control + smooth surface detection + pose correction
	\item \name: PID control + smooth surface detection + pose correction + half-precision floating point
\end{itemize}

Figures~\ref{fig:err_incre_opt} and \ref{fig:time_incre_opt} show the ATE and
computation time respectively for the different controller configurations. The
PID controller by itself achieves much of the time savings but violates the error constraint for 5 out of 17 inputs. This is not
acceptable since nearly a third of the inputs fail to meet the error constraint.
When the smooth surface detection is incorporated into the controller, the
overall error performance is improved, but 4 inputs still fail to meet the
5~cm error constraint. Note that surface detection introduces a $\sim$15\%
computation time overhead compared to the PID controller.

When pose correction is incorporated, all the benchmarks meet the error
constraint. It is interesting to note that this reduces the overall computation time
compared to using only smooth surface detection.  The reason is that this
setting improves localization and mapping, so more approximation can be done safely,
leading to reduced computation time. Finally, when half-precision floating-point
arithmetic is added, the overall computation time is improved by $\sim$5\% at
the cost of slightly worse error. Note that for a few of the trajectories,
\name produces less ATE than "PID controller + Detection + Correction" does.
This is because \kfusion is highly non-linear, so the
reduction in precision may affect the results in a positive way.

\subsection{Effectiveness of Online Control}
\label{sec:eval:knob_activities}

The results in the previous section showed the effectiveness of \name for entire trajectories. To get a better sense of how online control in \name works, it is useful to visualize how knob settings change from the beginning to the end of a complete trajectory. Figure~\ref{fig:knob_activity} is such a visualization for the
\bench{mr1} trajectory, which is one of the most difficult trajectories in our
benchmark set. The PID controller, even with smooth surface
detection, fails to meet the error constraint. In Figure~\ref{fig:knob_activity},
the $x$ axis represents frame number in chronological order.

The black line in Figure~\ref{fig:knob_activity_perf} shows instantaneous
trajectory error (ITE) over time when \name is used for the entire trajectory.
Those ITEs are computed after the execution by comparing the reconstructed
trajectory with the ground truth, which is assumed to be the trajectory
reconstructed with the most accurate setting of \Kfusion.
Figure~\ref{fig:knob_activity_csr} and Figure~\ref{fig:knob_activity_trigger}
show the settings of the \knob{csr} knob and the activation of smooth surface
triggers and pose correction triggers respectively.

To demonstrate the effectiveness of \name, we show another configuration in
Figure~\ref{fig:knob_activity_perf} (represented by the gray line). In
this configuration, we switch \name to the PID controller after frame 500.
During frames 520--590, the camera encounters a smooth surface and the PID
controller cannot deal with it by itself. As shown
in Figure~\ref{fig:knob_activity_perf}, ITE increases dramatically (peak ITE is $\sim$100~cm and is too large to be plotted on this scale) and never comes down to an acceptable level.
This shows that the optimization techniques used in \name are critical for complex trajectories like \bench{mr1}.

Figure~\ref{fig:knob_activity_csr} shows the activity of the most important knob \knob{csr}. Frames 200 to 400 are relatively easy and speed is low, so \name tunes the knob all the way up to approximate the computation. For frames between 700 to 1200, ITE gets larger because the velocity of the agent increases. As a result, \knob{csr} is set to the most accurate value by \name in order to handle the drift in the trajectory.

This example shows that \name can control the knobs dynamically to exploit opportunities to save computation time and energy while ensuring that the localization error is within some reasonable bound.

\subsection{PID Controller vs. Hierarchical Step Controller}
\label{sec:eval:naive_comparison}

As discussed in Section~\ref{sec:eval:odroid_perf}, the energy saving of
\name is not as good as its saving in computation time. The potential reason
is that the PID controller in \name frequently changes knob \knob{csr},
causing certain data buffers within the SLAM algorithm to be scaled up and down frequently.

To explore this issue, we designed a variant of \name, with a \emph{hierarchical} step controller as an alternative to the PID controller. Instead of generating exact knob settings, the hierarchical step
controller only generates two signals: increase or decrease
approximation. Signals are generated by comparing velocity and average
velocity to predefined thresholds, and checking whether the velocity is ascending
or descending, similar to the PID controller.
Instead of tuning multiple knobs simultaneously,
a hierarchical step controller tunes only one knob one level at a time,
following the order of importance of the knobs discussed in Section~\ref{sec:knob_rank}.
Other optimizations like surface detection, pose correction, and reduced-precision floating-point computation
are the same in the design variant.

\begin{figure}[hbt]
	\centering
	\vspace*{-15pt}
	\hspace*{-20pt}
	\subfloat{\includegraphics[scale=0.6]{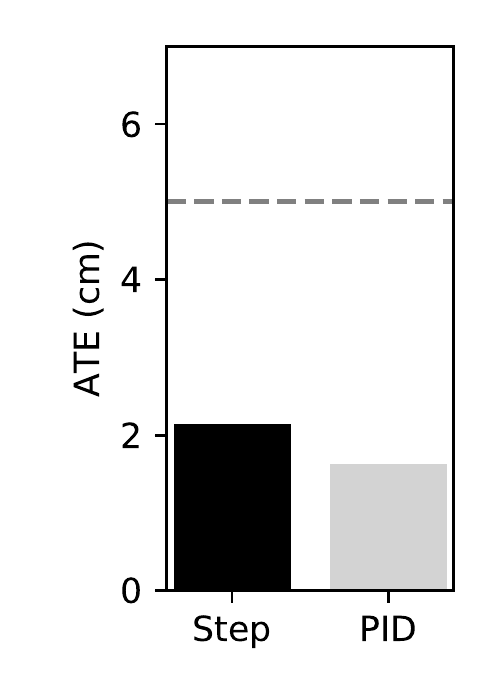}\label{fig:err_ctrl_comp}}
	\subfloat{\includegraphics[scale=0.6]{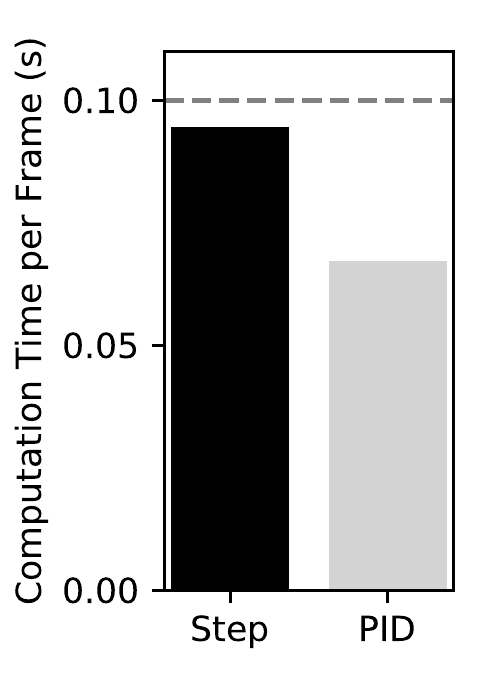}\label{fig:time_ctrl_comp}}
	\subfloat{\includegraphics[scale=0.6]{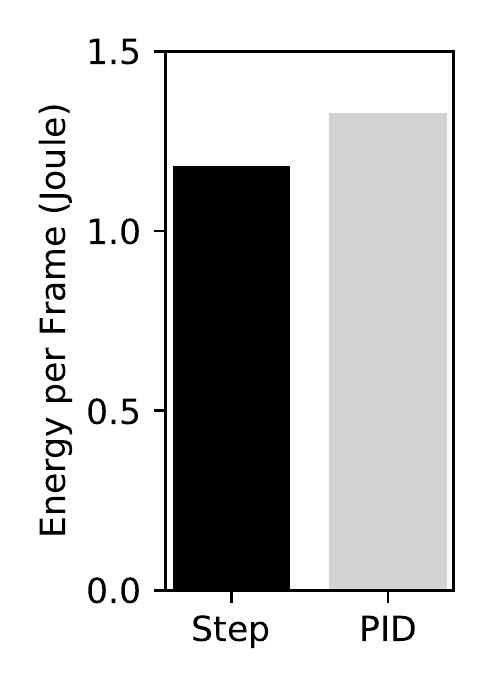}\label{fig:energy_ctrl_comp}}
	\vspace*{-10pt}
	\caption{\name with PID controller or hierarchical step controller.}

	\label{fig:naive_comp}
\end{figure}


Figure~\ref{fig:naive_comp} compares the performance of the two controllers.
Though the step controller version \name spends 29\% more
computation time than the PID version, it uses 12\% less energy. The
saving in energy can be explained by the fact that the tuning of the knob settings become smoother with the hierarchical step controller. However, the longer reaction time to get to ideal knob setting hurts its computation time.

\subsection{Comparison with Other Control Strategies}
\label{sec:eval:other_comparison}

This section compares \name with controllers based on two very different design philosophies: offline control and application-agnostic control.

\paragraph*{Comparison with offline control}
\label{sec:eval:offline_comparison}

We compare the performance of \name with offline control, given the best
offline knob setting of each trajectory. The best offline knob setting is
conducted by exhaustively searching the knob space described in Section~\ref{sec:knob_rank} for each input with the same error bound.
On average, \name is 1.32x faster than offline control. Interestingly, \name is
1.39x faster for real world inputs while the best offline knob setting is
slightly faster (3\%) for synthetic inputs (\bench{lr0, lr1, lr2}). The reason is that \name is
designed for general inputs whereas synthetic inputs are generally easier to process.

\paragraph*{Comparison with application-agnostic control}
\label{sec:eval:poet_comparison}

We compare \name with an application-agnostic control system for software
applications. Application-agnostic control systems have been used successfully
in the literature to control a diverse set of
applications~\cite{self-adaptive-software,poet-2015}. Next, we first briefly present a general strategy to design such a control system, which we call \emph{\textsc{TC}}, and then describe our adaptions to SLAM. 

\begin{itemize}

	\item The application execution is divided into intervals or windows
		  of some size (\eg, 32 or 64 frames in our experiments).
		  \textsc{TC} tries to meet the desired performance constraints and objectives on the average for each window.

	\item In each interval, TC tracks how well the performance
	      constraint has been met, and this information is used to decide
	      whether the system should be sped up or slowed down in the next
	      interval to better meet the performance constraint. This desired
	      performance level is normalized by the performance obtained by
	      setting knobs to their default values, and this dimensionless
	      quantity ("performance speedup") is used to find the knobs
	      for the next interval.

	\item To find knob settings for a desired performance speedup, TC consults a \emph{configuration table}, which returns
	      Pareto optimal knob settings for a given performance speedup (in
	      some cases, it returns a pair of knob configurations but this detail
	      can be ignored). The configuration table is constructed ahead of
	      time by profiling the program using representative inputs and knob
	      configurations.

\end{itemize}

We implemented an online controller for SLAM following the traditional
control design scheme described above. Instead of a performance speedup
requirement, each time interval is given an error budget (\ie, the
total amount of error allowed in the next interval) whose value is
computed using TC's strategy for determining
performance speedup. Since ground truth is not available for most of
our trajectories, we use velocity as a proxy for the actual error in
each interval (a reference velocity is defined as the required
velocity). The base velocity for each window, corresponding to the
base performance in the original traditional controller, is provided
by ground truth instead of implementing Kalman filtering~\cite{poet-2015}. Note that
this value is at least as accurate as what Kalman filtering estimates.
We used two approaches to build the configuration table. The
first one used the three synthetically-generated living room
trajectories \bench{lr0, lr1} and \bench{lr2} for profiling. Since
these trajectories are not representative of the entire set of
trajectories in our benchmark set, we would expect the performance of
the controller to be poor. The second approach is to use a more
diverse set of inputs, one from each scene in the benchmark set (for
example, \bench{mr1} is picked from the meeting room category).

\begin{table}[t]
  \renewcommand\sfsmaller{}
  \centering
  \newcommand{\n}{\ensuremath{n}\xspace}
  \newcommand{\f}{\ensuremath{\mathit{f}}\xspace}
  \renewcommand\k[1]{#1}
  \newcommand\num[1]{0.#1}
  \newcommand\snum[1]{0.0#1}
  \begin{footnotesize}
    \begin{tabular}{l|rr|rr}
      & \multicolumn{2}{c|}{\textbf{Living Rooms Config}} & \multicolumn{2}{c}{\textbf{Diverse Rooms Config}} \\
      \textbf{Ref} & \textbf{\# Error} & {\bf \# Better} & \textbf{\# Error} & {\bf \# Better} \\
      \textbf{Velocity} & \textbf{Violated} & {\bf Time} & \textbf{Violated} & {\bf Time} \\
      \midrule
      \bench{0.004} & 3.5 & 1 & 1 & 0 \\
      \bench{0.005} & 5.5 & 2 & 1 & 0 \\
      \bench{0.006} & 7 & 3 & 1 & 0 \\
      \bench{0.008} & 10 & 3 & 2 & 0 \\
      \bench{0.010} & 10 & 5 & 2 & 0 \\
      \bench{0.012} & 10 & 6 & 2 & 0 \\
    \end{tabular}
  \end{footnotesize}
  \caption{Traditional online controller performance}
  \vspace{-20pt}
  \label{tb:poet}
\end{table}

Table~\ref{tb:poet} shows the performance of the traditional control scheme compared to \name.
The evaluation includes running all the benchmarks using the traditional controller on each configuration table with different values for the reference velocity.
The \emph{Error Violated} column shows the number of inputs that violate the
5~cm error constraint, while the \emph{Better Time} column shows the number of
trajectories that satisfy the error constraint and have less computation time
than with \name. When the configuration table is built using only the living
room trajectories (\emph{Living Rooms Config}), the controller introduces
approximation too aggressively because the profiling set, living room
trajectories (\bench{lr0, lr1, lr2}), are relatively simple to approximate
comparing to other real world trajectories.
When the error budget (reference
velocity) gets looser, more trajectories achieve better computation time at the
cost of unacceptable tracking error. On the other hand, when profiling is done
with the diverse trajectory set (\emph{Diverse Rooms Config}), the error
constraint is rarely violated but the computation times are slower than with
\name because the  configuration table is overly conservative.

Prior work has shown that a traditional controller can control a diverse set of applications~\cite{poet-2015, self-adaptive-software}. However, high input sensitivity and low error tolerance characteristic of SLAM makes it difficult for a traditional controller to control \KFusion.
Intuitively, the configuration table is a model of system behavior that averages
over all the trajectories used in the profiling (training) phase, so the
controller cannot optimize the behavior of the system for the particular
trajectory of interest in a given execution. In addition, this controller does
not exploit the SLAM-specific techniques in \name such as smooth surface
detection and pose correction, which proved essential in Section~\ref{sec:eval:knob_activities}.

\section{Related Work}
\label{sec:relatedwork}

We discuss work on approximation in SLAM algorithms and on using control-theoretic approaches for optimal resource management.

\subsection{Approximating SLAM}

Recent work has used KinectFusion and the SLAMBench infrastructure to study
the performance impact of reduced-precision floating-point arithmetic in SLAM algorithms~\cite{wapco-2017, oh2016energy}. Unlike \name, these approaches do not exploit approximation at the algorithmic level.

\emph{Offline control} of KinectFusion has been explored by Bodin {\em et al.}\cite{bodin-2016} using an active learning technique. Given the entire trajectory ahead of time, they use a random forest of decision trees to characterize the input trajectory, and generate a Pareto optimal set of configurations that trades off computation time, energy consumption and the ATE. In addition to algorithmic knobs, they also explore approximation of compiler and hardware-level knobs. Their study is limited to a subset of frames for one synthetically-generated trajectory. Follow-up work utilizes the motion information of the autonomous agent to improve offline control and evaluates the offline approximation technique on other SLAM algorithms~\cite{hypermapper-tradeoff, slam-dse}. In contrast, \name performs online control so it does not need to know the entire trajectory before the agent begins to move. Adaptive control of knobs also permits \name to optimize knob settings dynamically to take advantage of diverse environments, which is not possible with offline control.

\subsection{Adaptive Resource Management}

Several systems have been proposed that aim to balance power or energy consumption
along with performance and program accuracy~\cite{jouleguard-sosp-2015,
meantime-atc-2016, poet-2015, khudia2015rumba}.
\emph{Rumba} is an online quality management system for detection and correction of large
approximation errors in an approximate accelerator-based computing environment.
Rumba applies lightweight checks during the execution to detect large approximation
errors and then fixes these errors by exact re-computation~\cite{khudia2015rumba}.
However, recomputation is not feasible in the context of SLAM algorithms because
it takes a stream of frame as input and the result of SLAM is used for real-time control. {\em JouleGuard} is a runtime control system that
coordinates approximate applications with system resource usage to provide
control-theoretic guarantees on energy consumption (\emph{i.e.}, will not exceed
a given threshold), while maximizing accuracy~\cite{jouleguard-sosp-2015}. The
control system uses reinforcement learning to identify the most energy-efficient
configuration, and uses an adaptive PID controller-like mechanism to maintain
application accuracy. \emph{POET} measures program progress and power
consumption, uses feedback control theory to ensure the timing goals are met,
and solves a linear optimization problem to select minimal energy resource
allocations based on a user-provided specification~\cite{poet-2015}. {\em
MeanTime} is a approximation system for embedded hardware that uses control
theory for resource allocation~\cite{meantime-atc-2016}. These techniques combine
PID-like control techniques to various parts of the system, and provide
empirical demonstrations of overall system behavior without taking domain
knowledge into consideration. 
Recent work on optimal resource allocation focuses on designing sophisticated
control systems using linear quadratic Gaussian control for
example~\cite{mimo-isca-2016,caloree-asplos-2018,spectr-asplos-2018}.
Unlike \name, they also do not exploit application-specific properties to perform control.


\section{Conclusion}
\label{sec:conclusion}

SLAM algorithms are increasingly being deployed in low-power systems but a
big obstacle to widespread adoption of SLAM is its computational expense. In
this work, we present \name, a PID control system, to approximate the
computation in KinectFusion, which is a popular dense SLAM algorithm. We show that the \name, augmented with insights from the application domain, is effective in reducing the computation time and the energy consumption, with acceptable bounds on the localization accuracy.


\iftoggle{includeAcks}{
  \section*{Acknowledgments}

This work was supported by NSF grants 1337281, 1406355, and 1618425, and by DARPA contracts FA8750-16-2-0004 and FA8650-15-C-7563. The authors would like to thank Behzad Boroujerdian for helpful discussions about SLAM.

}{}

\iftoggle{acmFormat}{
\clearpage
{
  \balance{}
  \bibliographystyle{ACM-Reference-Format}
  \bibliography{paper}
}
}
{
  \balance{}
  \bibliographystyle{IEEEtran}
  \bibliography{paper}
}

\end{document}